\def\year{2022}\relax
\documentclass[letterpaper]{article} % DO NOT CHANGE THIS
\usepackage{aaai22}  % DO NOT CHANGE THIS
\usepackage{times}  % DO NOT CHANGE THIS
\usepackage{helvet}  % DO NOT CHANGE THIS
\usepackage{courier}  % DO NOT CHANGE THIS
\usepackage[hyphens]{url}  % DO NOT CHANGE THIS
\usepackage{graphicx} % DO NOT CHANGE THIS
\usepackage{natbib}  % DO NOT CHANGE THIS AND DO NOT ADD ANY OPTIONS TO IT
\usepackage{caption} % DO NOT CHANGE THIS AND DO NOT ADD ANY OPTIONS TO IT
\DeclareCaptionStyle{ruled}{labelfont=normalfont,labelsep=colon,strut=off} % DO NOT CHANGE THIS
\usepackage{algorithm}
\usepackage{algorithmic}
\usepackage{latexsym}
\usepackage{multirow}
\usepackage{array, boldline, makecell, booktabs}
\usepackage{newfloat}
\usepackage{listings}
\floatstyle{ruled}
\newfloat{listing}{tb}{lst}{}
\floatname{listing}{Listing}
\title{Insta-VAX: A Multimodal Benchmark for Anti-Vaccine and Misinformation Posts Detection on Social Media}
\author{Mingyang Zhou$^{\thanks{Equal Contribution}1}$, Mahasweta Chakraborti$^{\footnotemark[1]1}$, Sijia Qian$^1$, Zhou Yu$^2$, Jingwen Zhang$^1$ \\
$^1$University of California, Davis \\
$^2$Columbia University \\
{\tt\small \{minzhou, mchakraborti, sjqian, jwzzhang\}@ucdavis.edu} \\
{\tt\small \{zy2461\}@columbia.edu}
}

\begin{document}

\maketitle

\begin{abstract}
Sharing of anti-vaccine posts on social media, including misinformation posts, has been shown to create confusion and reduce the public’s confidence in vaccines, leading to vaccine hesitancy and resistance. Recent years have witnessed the fast rise of such anti-vaccine posts in a variety of linguistic and visual forms in online networks, posing a great challenge for effective content moderation and tracking. Extending previous work on leveraging textual information to understand vaccine information, this paper presents Insta-VAX, a new multi-modal dataset consisting of a sample of 64,957 Instagram posts related to human vaccines. We applied a crowdsourced annotation procedure 
verified by two trained expert judges to this dataset. We then bench-marked several state-of-the-art NLP and computer vision classifiers to detect whether the posts show anti-vaccine attitude and whether they contain misinformation. Extensive experiments and analyses demonstrate the multimodal models can classify the posts more accurately than the uni-modal models, but still need improvement especially on visual context understanding and external knowledge cooperation. The dataset and classifiers contribute to monitoring and tracking of vaccine discussions for social scientific and public health efforts in combating the problem of vaccine misinformation. 
\end{abstract}

\section{Introduction}
Changes in the media landscape, especially with the rise of social media, have accelerated the creation and spread of disinformation and misinformation on a global level\cite{Vosoughi1146, d887a53478cf485287e60d6c92200a9b}. Exposure to anti-vaccine sentiments or associated vaccine misinformation can form false beliefs, arouse negative emotions, and reduce pro-vaccine attitudes, leading to vaccine hesitancy\cite{Betsch2011TheIO, Featherstone2020FeelingAT}.

On one hand, anti-vaccine information can simply signal an opinion that is against vaccines. On the other hand, vaccine misinformation is defined more clearly as ``a broad category of claims that inadvertently draws conclusions based on wrong or incomplete information''\cite{SOUTHWELL2019282} in the context of discussing vaccines. Vaccine misinformation thus specifically refers to false or inaccurate information judged by the scientific community's consensus contemporaneous with the time period\cite{Tan2015ExposureTH}. For instance, a personal anti-vaccine narrative that shows negative emotions toward vaccines does not necessarily contain misinformation. On the contrary, vaccine misinformation contains explicit or implicit claims or arguments based on false or incomplete information that argue about the scientific research, development and production, safety, or effectiveness of vaccines (e.g., the claim that vaccines cause Autism in children or the claim that vaccines are developed for population control). Although past research has shown many anti-vaccine messages circulated online contains misinformation claims\cite{DUNN20173033}, it is crucial to conceptually and operationally differentiate these two categories. If social media is ought to fact-check and tag misinformation, any automated approaches need to clearly set the boundaries that distinguish opinions from misinformation claims. While facing the challenges of vaccine hesitancy and recurring pandemic threats, accurately and efficiently identifying vaccine misinformation on social media while preserving personal free opinion expressions is a core issue to be tackled by the computational social science community. So far, past literature in NLP has not distinguished the two categories well and some have oversimplified anti-vaccine opinions as medical misinformation \cite{9253994}. 

The absence of tracking and moderating vaccine misinformation has left individuals vulnerable to strong persuasive agendas. Given limited attention and a lack of expertise, a layperson may be influenced by the wrong information without the capacity to assess the credibility and accuracy of its claims. Accordingly, recent efforts have focused on empowering individuals through developing algorithms that fact-check online information and provide related resources \cite{ZHANG2021106408}. A number of recent research has used multiple content analytical approaches to quantify anti-vaccine stance on social media. For example, Broniatowski et al.\cite{Broniatowski2020FacebookPT} categorized 204 Facebook pages and tracked the changes in framing of vaccination opposition over 10 years. Guidry et al.\cite{doi:10.2105/AJPH.2020.305827} analyzed 1000 Pinterest posts about HPV vaccination and found visual cues (e.g., a large needle) associated with anti-vaccine contents.  

Although many studies have extracted thematic topics in anti-vaccine information, few made efforts to distinguish anti-vaccine posts and misinformation posts. In addition, the majority of previous work focused on textual information. As contents on social media are increasingly relying on visuals to frame or supplement arguments and opinions, understanding the visual features associated with anti-vaccine information and misinformation is of great importance to enrich our classification of anti-vaccination communications on social media\cite{SELTZER2017170, 9253994}. 

To address the gaps, this paper presents a curated large-scale dataset of 64,957 Instagram posts relevant to vaccine discussions . We benchmark several state-of-the-art (SOTA) machine learning models by using different modalities to classify vaccine posts on dimensions of vaccine attitude (i.e., anti-vaccine or none) and vaccine misinformation (i.e., misinformation or none) based on an annotated sample of 5000 posts. The rationale for focusing on binary classifiers is that when considering the downstream impacts of the social media information, it is anti-vaccine and misinformation posts that have significant negative impacts on public health \cite{SOUTHWELL2019282}. Experiments show that multimodality is helpful to improve the detection accuracy, but the help from visual cues is only incremental due to the challenges of visual context understanding. We hope our findings point to more effective directions for multi-modal machine learning methods to address vaccine misinformation content moderation. We provide insights into the different visual and textual cues that are highly correlated with anti-vaccine attitude and misinformation to help the general audience to identify such social media content. 

\section{Related Work}
\subsection{Anti-Vaccine and Vaccine Misinformation Posts}
Anti-vaccine discourses and vaccine misinformation encompass multiple and complex dimensions of reasoning and emotions, with explicit arguments or implicit suggestions involving references to nuanced social contexts. Prominent explicit arguments include vaccines and their ingredients cause illnesses and adverse effects \cite{Guidry2015OnPA}. These beliefs are supported by a generic frame that argues the scientific community cannot be trusted because of inherent uncertainties or potential malpractices \cite{Zimmerman2005VaccineCO}. In addition, personal narratives eliciting strong emotional resonances with the viewers are also typical types of messages featured in such anti-vaccine discourses where stories featuring children injured from vaccines are common examples \cite{Margolis2019StoriesAH}. The second prominent type focused on the public's declining trust in authorities and prompts conspiracy ideation regarding the vaccination programs, speculating on how governments and pharmaceutical companies cover up the ``truth of vaccines'' from the public \cite{Davies22}. Lastly, the discussions on vaccines can also invoke religious or policy oppositions centering on the ethical grounds of vaccine mandates. Although this type of discussions does not directly touch upon calculations of vaccine benefits and risks about specific diseases, it can arouse strong political and emotional reactions that can negatively impact favorable opinions toward vaccines \cite{Haber2007TheHV}.

\begin{figure}[h!]
\centering
\includegraphics[width=7.5cm]{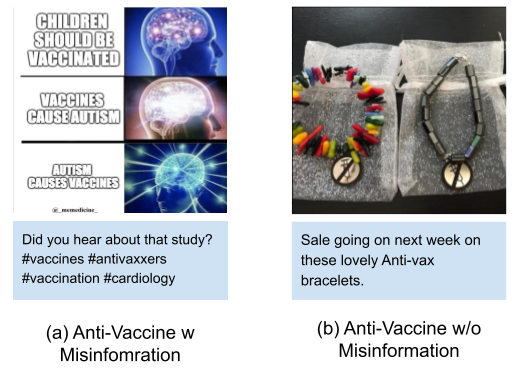}
\caption{Sample anti-vaccine Instagram posts that are labeled to having misinformation versus not having misinformation.}
\label{fig:fig-DIF}
\end{figure}

According to the definition of vaccine misinformation (i.e., false or inaccurate information judged by scientific consensus) and based on previous literature operationalizing vaccine misinformatio \cite{BURKI2019e258, SOUTHWELL2019282}, we deem explicit claims and arguments for lack of vaccine necessity, ineffectiveness, lack of safety, danger, unknown harms of vaccines, and vaccine conspiracies as misinformation because they are false claims inconsistent with established expert consensus and cumulative scientific evidences. Vaccine conspiracies are deemed as misinformation because they draw conclusions based on pure speculations or false and incomplete information. Other than these, we deem personal stories or expressive opinions that do not involve claims as non-misinformation but as indicating attitudes toward vaccines. Figure~\ref{fig:fig-DIF} provides two example posts, with (a) labeled as both anti-vaccine and containing misinformation (i.e., claiming vaccines cause autism) whereas (b) labeled as anti-vaccine without presence of misinformation (i.e., just showing anti-vax bracelets). 

\subsection{Misinformation Detection}
Previous research on misinformation detection has utilized a variety of approaches, including analyzing content features\cite{Ito2015AssessmentOT,Hu2014SocialSD}, social context features \cite{DBLP:journals/corr/abs-1708-01967, Volkova2018MisleadingOF, Tacchini2017SomeLI}, and leveraging knowledge graphs \cite{Ciampaglia2015ComputationalFC} and the wisdom of the crowds\cite{Qazvinian2011RumorHI}. Recently, deep learning-based methods that abstract a high-level representation of data has been often adopted to conduct misinformation detection \cite{Li2018ARE, Yu2019AttentionbasedCA, Dong2019MultipleRS}.

However, given the complexity of anti-vaccine and misinformation contents, effective machine-learning models that can understand and extract useful textual and visual signals from social media posts on vaccine attitude and misinformation are still not well developed. Only few studies have made an effort to tackle this problem. \cite{Kostkova2016VACMA} developed an interactive dashboard integrating social media contents on Twitter and news coverage from the monitoring tool MediSys \cite{Ralf2008MedISysAM}, showing public debate related to vaccines. This platform helps experts to track public understanding of vaccines. However, it does not directly identify misinformation. Most relevant to our work, Zuhui et al.\cite{9253994} proposes a multi-modal ensemble model to detect anti-vaccine posts from 30K Instagram posts. However, it does not distinguish misinformation from anti-vaccine posts. It also does not delve into analyzing the specific textual and visual characteristics correlated with anti-vaccine posts, which we deem as of great value for the NLP community and the public to further understanding the vaccine discourses online. 

The innovation of our paper is twofold. First, our annotations and classifiers distinguished misinformation from anti-vaccine attitude in Instagram posts. Second, we present experiments with several SOTA machine learning models and point out the challenges of employing different modalities. With our findings, we hope to spur future research to build strong anti-vaccine and misinformation classifiers by better leveraging the multimodal context.

\section{Insta-VAX Dataset}
\subsection{Data Collection}
To collect Instagram posts that express diverse opinions toward vaccines, we carefully curated 39 Instagram hashtags that are related to vaccines, shown in Appendix. From the 39 hashtags, we collected 494,249 posts.\footnote{We used the python-based Instaloader package as the data scraper.} To further clean the dataset, we deleted the posts about animal vaccines via keyword matching. We used a large corpus of keywords
\footnote{The keywords were collected categorically from Google search trends and recommendations. For a complete list, please refer to Appendix} 
that covered pet food brands, stores, rescue organizations, vet diseases, and breeds, etc. Posts with multiple mentions of the keywords from the list were excluded. We also filtered out posts whose captions contained non-English texts detected by Google Translate API. As a result, we obtained a total sample of 64,957 posts published between June 2011 and April 2019.

\paragraph{Ethics}
In line with Instagram data policies \cite{instagram_data_policy} as well as user privacy we only gathered publicly available data with target hashtags that is obtainable from Instagram. 

\subsection{Crowdsourced and Expert Annotations}

We combined crowdsourcing and expert annotation approaches to label the posts. At the first stage, we used Amazon Mechanical Turk (MTurk) for crowdsourcing. We selected a random sample of 4,997 posts, and split them further into 100 batches of 50 posts, allowing four independent workers to work on each batch. Workers were asked to label the post's attitude towards vaccine and whether the post has misinformation claims about vaccines in each post based on both images and captions. Workers were provided with scholarly definitions of anti-vaccine attitudes and vaccine misinformation claims. 
% Appendix. \ref{insta-interface} presents the annotation interface with instructions for workers. 
Specifically, we asked two questions: 1 ``What is the attitude of the post?'' (response choices included `anti-vaccine,' `pro-vaccine,' and `neutral') and 2 ``Does the displayed post contain misinformation?'' (`yes,' `no,' and `not sure'). Responses were recorded as binary inputs as anti-vaccine vs. none and misinformation vs. none. Among the 5,000 posts, three did not retain original images and thus were eliminated from further analyses, resulting in 4997 posts for annotation. Each worker was paid 2 dollars for annotating 50 posts.

After receiving the crowdsourced labels, we trained two expert coders, one undergraduate and one graduate student, to examine and finalize the coding of vaccine attitude and misinformation claims for all the posts. The two coders were trained by a senior researcher specializing in public health and vaccine misinformation communication. Specifically, the two coders examined scientific and organizational webpages (i.e., CDC, FDA, HHS) on common vaccine misinformation claims. 

Each coder independently examined a randomly selected sample of 100 posts and established high inter-coder reliability (anti-vaccine attitudes: Cohen’s Kappa = $0.82$; misinformation: Cohen’s Kappa = $0.78$). The two coders discussed and resolved all disagreements. Then the two coders proceeded to check the labeled posts from MTurk, with each independently checking half of the sample. We examined the crowdsourced results by the criteria of majority votes, which is defined as receiving the same vote from three out of four workers on the post. For question 1 regarding vaccine attitude, $84.9\%$ posts received clear majority votes whereas question 2 regarding misinformation received $75.8\%$ majority votes. For the posts receiving majority votes, the two coders checked and verified if they were correct. For the posts that received split votes, the coders provided final judgements. Thus, we determined the ground truth basing on both crowdsourced votes and expert judgements, and we used the finalized labels to train the models.  

\begin{table}[h]
\centering
\small
\begin{tabular}{|l|l|}
\hline
\textbf{Total posts}                         & 64,957          \\ \hline
\textbf{Labeled posts}                       & 4,997             \\ \hline
\textbf{Anti-vaccine posts}                  & 1,373             \\ \hline
\textbf{Misinformation posts}                & 1,115               \\ \hline
\textbf{Misinformation \& Anti-vaccine Posts} & 1,101 \\ \hline
% \textbf{} & 768               \\ \hline
\textbf{Training set}                      & 4,000 \\ \hline
\textbf{Testing set}                       & 997 \\ \hline
% \textbf{Time span}                             & 06/2011 - 04/2019 \\ \hline
% \textbf{Total word counts}                      & 294,388           \\ \hline
\textbf{Avg. words per post}           & 55.27             \\ \hline
% \textbf{Total word counts (labeled)}                  & 54,461            \\ \hline
\textbf{Avg. words per post (labeled)}       & 51                \\ \hline
\end{tabular}
\caption{Insta-VAX dataset statistics}
\label{tab:tab-stats}
\end{table}
\subsection{Dataset Details}
In table.\ref{tab:tab-stats}, we provide an overview of the statistics of our dataset. There is a clear overlap between the posts that contain misinformation and that express anti-vaccine attitudes, with $98.7\%$ misinformation posts containing anti-vaccine attitudes. However, only $80.2\%$ posts that express anti-vaccine attitudes contain misinformation. This shows the possibility of differentiating the two content categories challenges the simple assumption with previous research \cite{9253994}  which collapse the two. The annotated sample was then split into $80\%$ for training, and $20\%$ for testing using stratified split. 

section{Experiment}
We set up baseline machine learning models, which leverage different inputs from our datasets to detect anti-vaccine posts and misinformation posts. By analyzing the limitations of the baseline models, we hope to provide insightful suggestions for future work on multi-modal misinformation detection using datasets similar to Instagram posts.

\subsection{Models}
We experimented with three SOTA uni-modal models and multi-modal models to detect anti-vaccine and misinformation posts. Specifically, the uni-modal models either use the images or the texts to classify a post, while the multi-modal models use both texts and images as the inputs.  All the output layers for classification are multi-layer perceptrons followed by a softmax layer. 

\paragraph{Text-Only Uni-modal Models}
First, we mainly consider the state-of-the-art transformer-based pre-trained language model BERT \cite{devlin-etal-2019-bert}. Given the post text, we tokenize the input text into WordPieces \cite{WordPieces} and feed it into BERT. BERT then converts the text tokens into embedding vectors and process the vectors through a multi-layer transformer architecture \cite{NIPS2017_3f5ee243} to learn contextualized embeddings of the input text. Finally, we extract the representation of $\textit{[CLS]}$ token as the global representation of the input text and feed it into the output layer to classify the text. 
% For our experiment, we utilize the pre-trained BERT-Based, Uncased model and fine-tune it on our dataset with or without conducting masked language modeling pre-training on the raw instagram post. 
\paragraph{Image-Only Uni-modal Models}
For Image-based Unimodal Models, we employ the convolutional neural networks, ResNet-152 \cite{He2016DeepRL}. Given an input image, we process the image through ResNet-152 and extract the average pooling as output, which yields a 2048-dimensional vector for each image. The extracted vector is then forwarded through the output layer to perform classification. 

\paragraph{Multi-Modal Models}
For multi-modal models, we experiment with the state-of-the-art method UNITER \cite{UNITER}. UNITER is an extensions of the BERT model that takes the concatenation of image and text and fuses the two modalities through a single stream multi-layer transformer to extract the joint representation. The joint representation is then forwarded through the output layer to classify the multi-modal context into different labels. UNITER encodes the input image as a set of  visual features for the detected region of interests (ROI) from a pre-trained object detector \cite{BUTD}. 
% UNITER is an extension of BERT model that learns the universal representation of vision and language through weakly supervised learning from large corpus of image captioning datasets. 
% Given a pair of image and text, the visual representation is extracted as a set of image regions from a pre-trained object detector \cite{BUTD}, and the language representation is a list of sub-word tokens obtained from WordPiece \cite{WordPieces}.  

\begin{table*}[h]
\centering
\begin{tabular}{c|c|c|c|cc}
\Xhline{1pt}
\textbf{Modalities}          & \textbf{Models}         & \textbf{Text Src} & \textbf{Raw Finetune}                 & \textbf{anti-vaccine} & 
\textbf{misinformation} \\ \Xhline{1pt}

-                            & Majority Vote       & -      & -  & $42.3\%$              & $43.8\%$                                \\ \Xhline{1pt}
Image                   & ResNet-152              & -              &  -                 &  $66.9\%$              & $69.3\%$                             \\ \Xhline{1pt}
\multirow{3}{*}{Text}                    & \multirow{3}{*}{BERT}                    & caption                 & -                & $84.5\%$                 & $79.9\%$                                  \\ 
 & & caption + OCR &  - &  $84.6\%$ & $81.3\%$ \\ 
 & & caption & MLM & $85.5\%$ & $82.5\%$ \\
 & & caption + OCR &  MLM &  $87.9\%$ & $85.1\%$ \\ \Xhline{1pt}
 \multirow{3}{*}{Image+Text}                    & \multirow{3}{*}{UNITER}                    & caption                 & -                & $85.6\%$                 & $82.0\%$                                  \\ 
 & & caption + OCR &  - &  $86.2\%$ & $82.5\%$ \\ 
 & & caption & MLM & $87.7\%$ & $85.0\%$ \\
 & & caption + OCR &  MLM &  $\textbf{89.1\%}$ & $\textbf{87.3\%}$ \\ \Xhline{1pt}
%  & & caption + OCR &  MLM+MRC &  $88.4\%$ & $86.4\%$ \\ 
% \multirow{2}{*}{Multi-Modal} & \multirow{2}{*}{UNITER} & w/o. weakly supervised finetuning & $84.1\%$                 & $78.4\%$                                  \\ \cline{3-5} 
%                              &                         & w weakly supervised finetuning    & $86.6\%$                & $83.0\%$                                 \\ \hline
\end{tabular}
\caption{Results on detection posts containing anti-vaccine sentiment or misinformation from methods with different combination of input modalities. }
\label{tab:tab-result}
\end{table*}
\subsection{Experiment Settings}
Given the image and the caption of an Instagram post, we evaluate the three proposed methods on two binary classification tasks: (1) what is the attitude of the Instagram post toward vaccine (2) whether the Instagram post contains misinformation. By comparing the performance of the three proposed methods that leverages input from different modalities, we can understand how each modality is contributing to help the model to detect the anti-vaccine sentiment and the misinformation content. As our dataset is highly imbalanced, we use macro-F1 instead of accuracy as the evaluation metric.
During the fine-tuning stage our data, we also experiment with two variances. First, as the images of the Instagram are often memes, we also extract the embedded text (OCR) from the images with Google Vision API as an additional resource to study the contribution from vision. Next, for the transfomer-based methods like UNITER and BERT, we experiment both settings including and excluding of performing weakly-supervised learning with their proposed pre-training objectives on our un-annotated raw data. In our experiment, we apply masked language modeling, where the model learns to predict a masked word from the surrounding context. For every experiment, we always run 3 times with different random seeds and report the average results of the all 3 rounds. 
% we also experiment inclusing and exclusion of various of their proposed pre-training objectives including masked language modeling (MLM) and masked region classification (MRC) 

\subsection{Training Details}
We initialize Transformer-based models such as BERT and UNITER with pre-trained BERT-based, uncased model weights. For image-only uni-modal model, we initialize Resnet-152 with the pre-trained weights obtained from the ImageNet Classification task \cite{imagenet_cvpr09}. All parameters of the transfomer-based models are optimizable. However, for image-only uni-modal model, we only finetune the last ResNet block and the final fully connected layer to maintain the consistency of the early level features. For all models, we use Adam optimizer \cite{Kingma2015AdamAM} with a linear warm-up for the first $5\%$ of training and set the learning rate to $5e-5$. The learning rate is decayed linearly proportional to the training steps. Since the dataset labels are not balanced, we weigh the class labels by their inverse frequency during optimization. Fine-tuning is conducted on 2 Nvidia RTX-2080 Ti GPUs for 5 epochs. When pre-train BERT and UNITER on the raw Instagram posts, train the model for 10 epochs on 2 Nvidia RTX-2080 Ti GPUs. For the weakly supervised learning, we also use Adam Optimizer with the learning rate set to $5e-5$ to optimize the model.
\begin{figure*}[h!]
\centering
\includegraphics[width=15cm]{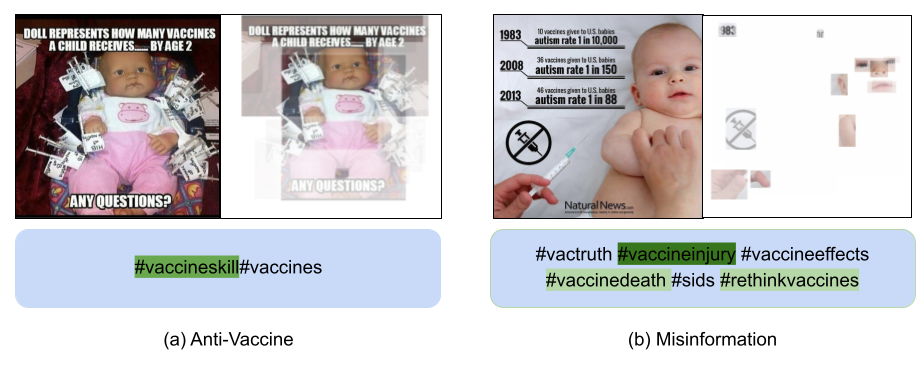}
\caption{Plot of the attention map on image and the caption on the Instagram posts on samples of the anti-vaccine Instagram post and misinformation post where the model predict correctly.}
\label{fig:fig-attention}
\end{figure*}

\section{Results and Analysis}
\subsection{Quantitative Results}
Table.\ref{tab:tab-result} summarizes the model performances on our test set for the task of detecting posts that express anti-vaccine attitudes and posts that contain misinformation. Both image-only baseline and text-only baseline have significantly outperformed the majority vote baseline, demonstrating that both modalities are useful for models to learn the appropriate knowledge to perform the task. However, the text resource is clearly more helpful than the visual context for both classification tasks, given that the BERT model achieves over $12\%$ classification accuracy against the image-only baseline. Despite that, UNITER, which takes input from both modalities, achieves the best performance. The help from vision context is still quite limited as the improvement over BERT is relatively incremental. One possible reason is that many images on Instagram posts are memes containing dense OCR texts but not many meaningful visual objects. By extracting the OCR texts from the image, and encoding it as part of the text input, we can observe clear improvement on both BERT and UNITER. 
We also observe that the transformer-based models significantly benefit from the weakly supervised fine-tuning on the raw Instagram posts. When fine-tuning BERT and UNITER with masked language modeling, they both gain over $2\%$ improvement on anti-vaccine detection and over $3\%$ on misinformation detection. We hypothesize that such pre-training objectives can alleviate the severe domain shift between the original pre-training corpus used by BERT and our vaccine centered Instagram data. 

% We include the visualization of the attention on multi-modal contexts predicted by the model in Appendix~\ref{attention-visuali}. We find that the model often attend to common objects instead of the salient objects associated with vaccine or medical related items. This is due to the bias in the visual representation of UNITER. Future work needs to build image encoder that has less reliance on the supervision from a pre-training corpus.
% When the UNITER is fine-tuned on raw data with both masked language modealing and masked region classification objectives, the classification accuracy is reduced compared to the model that is fine-tuned with just masked-language-modeling.  

\section{Attention Visualization} \label{attention-visualization}
To understand how the model interprets the multi-modal context when it makes the prediction result, we visualize the attention predicted by the model on the image and text input when our model successfully detects the posts. Some samples are presented in Figure~\ref{fig:fig-attention}. The model can attend to the salient text information that has strong indication to either anti-vaccine sentiment or misinformation, such as the hashtags `$\#\textit{vaccineskill}$', `$\#\textit{vaccineinjury}$', and `$\#\textit{rethink vaccines}$'. In comparison to the accurate attention to the texts, the model struggles to attend to critical objects that may be correlated to anti-vaccine sentiment. In both examples, the model attends more to the baby face instead of attending to the needles eliciting negative thoughts or emotions like fear. Especially on the misinformation example, the model concentrates on the salient local parts of the baby's face. We hypothesize the failure to attend to the more relevant image regions comes from the bias in the visual representation of UNITER, which is more prone to common objects observed in their original training datasets. Future work needs to build image encoder mechanisms that have less reliance on the supervision from a pre-training corpus, such as the grid-based image feature extractor employed by MMBT \cite{mmbt}.

\begin{figure*}[h!]
\centering
\includegraphics[width=16cm]{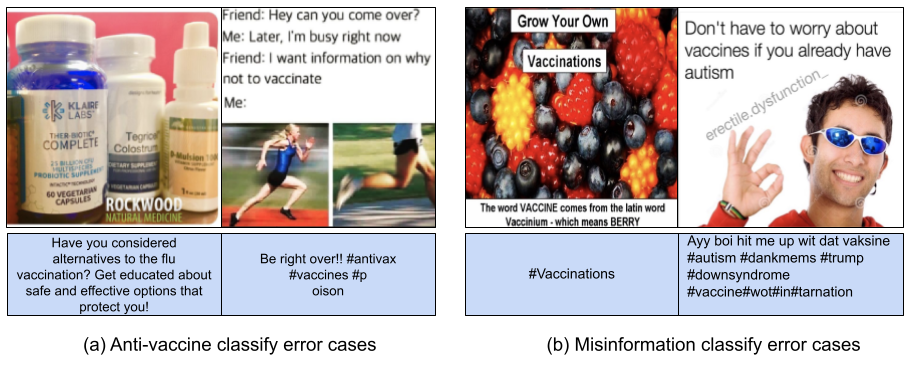}
\caption{Example of testing failure cases when applying the best multi-modal model. On the left-hand side are the two failed samples of the anti-vaccine posts, and on the right-hand side are the two failed samples of the misinformation posts. The captions of each image is displayed in the blue box under each image.}
\label{fig:fig-EA}
\end{figure*}
\subsection{Ablation Study on Hashtags}
Hashtags are important segments in instagram captions, which often contain keywords that indicate the attitude towards vaccines. To avoid the model achieving high accuracy by capturing only the mapping between critical hashtags and anti-vaccine sentiments or misinformation, we conduct an ablation study to train UNITER on the dataset with only hashtags or the captions without hashtags for the anti-vaccine data to verify our datasets' validity. With the training on just captions without hashtags, we achieve $76.2\%$ F1 score, while we achieve $78.1\%$ F1 score with training on just hashtags. Both are significantly lower than $84.5\%$ F1 score, as reported in table~\ref{tab:tab-result}, when UNITER is trained with all information in the captions. This indicates that both hashtags and the rest of the captions contribute critical signals for detecting anti-vaccine post. Looking into the error cases resulted from the model trained on hashtags only, we find that a critical portion of non-anti-vaccine posts contains anti-vaccine hashtags for satire purposes. In these cases, the model which considers information other than hashtags in the captions often achieves better predictions.
We intend to understand whether other context outside hashtags would still contribute significant cues to detect anti-vaccine vaccine and misinformation. 

\subsection{Error Analysis}
To further understand the complexity of social media vaccine posts, we examine samples from the test set that our best model UNITER predicts incorrectly. Such examples are shown in Figure~\ref{fig:fig-EA}. Overall, our best model can detect the post that contains anti-vaccine attitudes more accurately, in comparison to detecting posts that contain misinformation, with prediction accuracy of $79\%$ and $73\%$ respectively. 

The majority of anti-vaccine posts that are predicted correctly by our model contain key phrases or hashtags that express strong negative sentiments toward vaccines, such as `$\# \textit{antivax}$', and `$\#\textit{vaccinetoxic}$'. Our model experiences a hard time classifying posts with ambiguous, suggestive, or sarcastic expressions. Examples are shown in Figure~\ref{fig:fig-EA}, where the images contain vaccine-irrelevant objects and the captions use suggestive arguments. The full meaning is only interpretive when combing texts and images together. Errors on these posts suggest future multi-modal models need to build better mechanisms to extract visual representation and ground such visual information into its associated textual context.

As posts containing misinformation are highly correlated with posts that express anti-vaccine attitudes, a large portion of the failure detection cases are also shared. Example cases shown in Figure~\ref{fig:fig-EA} do not contain clear false claims in captions or hashtags and the images are also seemingly irrelevant. Correct identification on the misinformation about vaccines and autism requires external knowledge. As shown in Table~\ref{tab:tab-result}, performing weakly supervised learning on the corpus related to this topic would dramatically help the model to obtain such knowledge to detect misinformation, which helps the multi-modal model to gain over $5\%$ detection accuracy. However, to accurately detect the misinformation, background fact-checking and learning with domain specific knowledge needs to be incorporated.

\begin{figure}[h!]
\centering
\includegraphics[width=\linewidth]{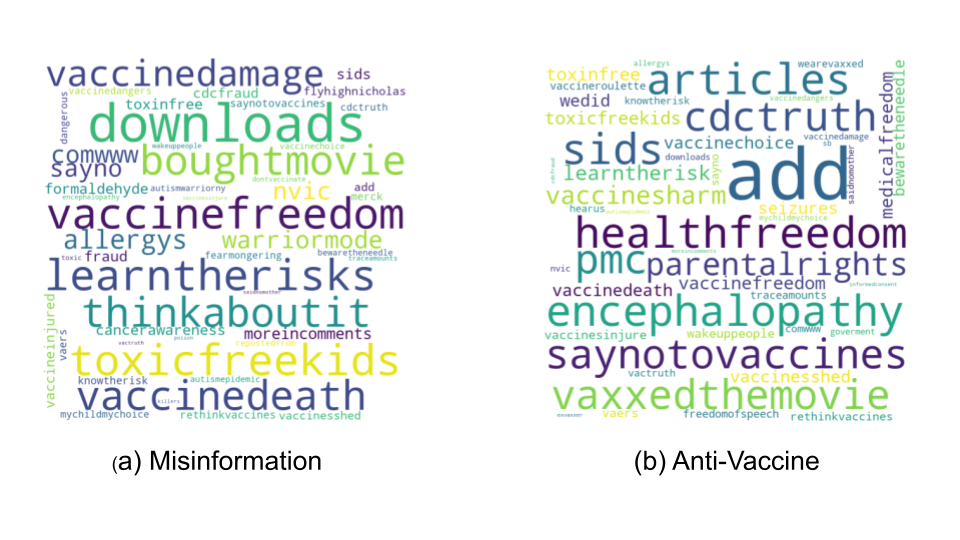}
\caption{The word cloud of the most salient words that are highly correlated with anti-vaccine and misinformation from the Instagram posts. The font of each word is proportional to the important score for each word.}
\label{fig:fig-wordcloud}
\end{figure}
\section{Textual and Visual Characteristics}
Given the above-mentioned limitations, we conduct further analyses to identify textual and visual characteristics that are prominent in anti-vaccine posts and misinformation posts. These insights could be used to build more accurate models to move forward research on social media vaccine content moderation.

\subsection{Textual Cues and Trends} 
To identify textual characteristics for anti-vaccine and misinformation posts, we compute the relative importance of the words from the Instagram captions and the embedded texts that are correlated with the anti-vaccine and the misinformation posts. We design a score function to compute the word importance as:
\begin{equation}
  score(w) = y_{w} / \hat{y}_{w}
\end{equation}
 where $y_w$ and $\hat{y}_w$ stands for the portion of texts that contain $w$ from the posts belonging to class $y$ and the opposite class $\hat{y}$. Similar to TF-IDF, it values the high frequency of the word in the post belonging to the target class but penalizes the word that is too common to appear in all posts from the corpus. Words that appeared less than 5 times are filtered to prevent favoring low-frequency rare words.

We display the most important words for anti-vaccine and misinformation posts as a word cloud in Figure~\ref{fig:fig-wordcloud}, where the size of the word is proportional to its importance score. Hashtags, for example `$\#\textit{vaccinefreedom}$' and `$\#\textit{saynotovaccines}$', that directly indicate negative attitudes toward vaccines clearly stand out in both misinformation and anti-vaccine posts. Besides the negative sentiment to vaccines, misinformation posts also have a clear emphasis on the risk of vaccines with keywords such as `allergies,' `vaccinedeath,' and `learntherisk.' It is highly likely that the content that discusses the vaccine risk conveys misinformation.  From the keywords `toxicfreekids' and `parentalrights,' we can also tell that the anti-vaccine and misinformation contents often target to invoke the issue with vaccination rights and to impact parents with young kids who might be more vulnerable to become vaccine-hesitant.

% The score to evaluate the importance of each word values the high frequency of the appearance in the post that is anti-vaccine or misinformation 
% value two factors: (1) High frequency of the  appearance in the post that is anti-vaccine or misinformation. (2) Low frequency of apperance in the post that belongs to the opposite category. 
 
% \begin{figure}[h!]
% \centering
% \includegraphics[width=\linewidth]{acl-ijcnlp2021-templates/Attitude_hsv.png}
% \caption{HSV histograms for Vaccine stance}
% \label{fig:fig-wordcloud}
% \end{figure}
\begin{table}[h!]
\small
\begin{tabular}{|c|l|}
\hline
\textbf{Task}  & \textbf{Most Correlated Objects}                              \\ \hline
Anti-Vaccine   & Food, Tie, Drink, Syringe, Container      \\ \hline
Misinformation & Suit, Syringe, Drink, Food, Container \\ \hline
\end{tabular}
\caption{The top-5 image objects that is highly correlated with the instagram posts with anti-vaccine sentiment or misinformation. }
\label{tab:tab-objects}
\end{table}
\subsection{Visual Cues and Trends}
In order to understand the visual characteristics across the two dimensions of anti-vaccine attitudes and misinformation, we analyze the correlation between the objects appeared in the images and the posts. We apply Google Vision API to detect the objects on each image, and in total we have 256 distinct object types detected in our dataset. Similar to finding the important words for each class, we also employ the importance score defined in the previous section to identify the salient objects that are highly correlated with anti-vaccine and misinformation. The top 5 objects with the highest importance scores for anti-vaccine posts and the misinformation posts are summarized in Table\ref{tab:tab-objects}

There is a big overlap between the salient objects that appear in the posts with anti-vaccine sentiment and the posts containing misinformation.  The frequent appearances of ``food'', ``drink'' , and ``container'' in anti-vaccine posts are in general consistent with some of their claims that lifestyles and holistic food habits can partially or wholly replace vaccination. The frequent reference to syringes in the anti-vaccine images is consistent with previous research suggesting anti-vaccine and misinformation posts often exploit the negative emotions such as fear \cite{Featherstone2020FeelingAT}, anger, and pain associated with needles to promote vaccine hesitancy.

\section{Conclusion and Future Work}
Our paper introduces a new multi-modal dataset, Insta-VAX, to detect anti-vaccine contents from the social media Instagram on the dimensions of vaccine attitudes and misinformation. We conceptually and operationally separate out the attitudinal posts from the misinformation posts and demonstrate that detecting misinformation is much more challenging and requires additional contextual knowledge. %We benchmark the proposed dataset against several SOTA baseline methods, including computer vision models, language models, and multi-modal models. 
Extensive experiments indicate that despite the generally superior performance of  multi-modal models in detecting anti-vaccine Instagram posts, text context still plays the major part; visual information only helps slightly because the pre-trained vision extractor has bias and is not trained for this particular domain-specific task. We also find detecting misinformation in vaccine posts more challenging because it requires the model to incorporate domain-specific knowledge and common sense beyond the dataset itself. %Finally we conduct deeper analysis on our data to gain insights on the visual cues and texture cues that can suggest the anti-vaccine sentiment or misinformation. 
In addition, we find that besides certain hashtags stand out to clearly indicate anti-vaccine sentiment, discussion on the risks of vaccines is potentially a red flag for misinformation. By analysing the objects from the images, we found healthy food, such as vegetables, and syringes are potential signals for anti-vaccine attitudes. 

In the future, we intend to continue improving the multi-modal model to detect anti-vaccine and vaccine misinformation from social media by building more dynamic visual feature extractor and mechanism to incorporate domain-specific knowledge and broader social political discourses about vaccine science, vaccination programs, and ethical concerns. We also want to explore transfer learning to generalize the knowledge we gained from the Instagram data to combat anti-vaccine content from other social media platforms.  

\newpage
\bibliography{aaai22}

\newpage
\appendix
\section{Instagram Hashtags} \label{insta-tags}
\begin{center}
\fbox{\begin{minipage}{20em}
\#antivaccine, \#provaccine, \#vaccineinjuryawareness, \#antivaccines, \#provax, \#vaccines, \#antivax, \#provaxx, \#vaccines\_facappts, \#antivaxmemes, \#provaxxer, \#vaccinesharm, \#antivaxxer, \#saynotovaccines, \#vaccineskil, \#dontvaccinate, \#vaccinate, \#vaccineskillkids, \#dontvaxx, \#vaccinated, \#vaccinessavebro, \#educatebeforeyouvaccinate, \#vaccinateyourkids, \#vaccinessavelives, \#measles, \#vaccination, \#vaccineswork, \#measlesoutbreak, \#vaccinations, \#vaccinesworks, \#novaccines, \#vaccine, \#vaccinetruth, \#provacc, \#vaccinefree, \#vax, \#provaccination, \#vaccineinjury, \#vaxxed \end{minipage}}
\end{center}

\section{Keywords for Animal Vaccine Posts Filtering} \label{animal-filtering}

\begin{table}[h!]
\small
\begin{tabular}{|l|l|}
\hline
\textbf{Keywords Type}                & \textbf{Keywords}                                                                                                                                                                                                                                                                                                               \\ \hline
\textbf{Pet Health}           & \begin{tabular}[c]{@{}l@{}}adenovirus, parainfluenza, spay, \\ neuter, parvovirus, petcare, deworm, \\ petcare, breed, spay, neuter, kennel, \\ rabies, rabid, fvrcp, distemper, \\ veterinarian, vet\end{tabular}                                                                                                              \\ \hline
\textbf{Pet Brands}           & \begin{tabular}[c]{@{}l@{}}dogwalker, iams, orijen, avoderm, \\ dogswell, merrick, spca, petco, \\ aspca, petsmart, shelter, rescue\end{tabular}                                                                                                                                                                                \\ \hline
\textbf{Pet Breeds}           & \begin{tabular}[c]{@{}l@{}}sphynx, retriever, labrador, bulldog, \\ beagle, poodle, rottweilers, terrier, \\ dobermann, chihuahua, pinscher, \\ dachshund, pedigree, pug, husky, \\ pomeranian, corgi, mastiff, shiba, \\ maltese, bichon, hound, dalmatian, \\ spaniel, birman, ragdoll, siames, \\ shitzu, akita\end{tabular} \\ \hline
\textbf{Animals} & \begin{tabular}[c]{@{}l@{}}animal, dog, pet, cat, pup, puppy, \\ canine, feline, cow, cattle, \\ livestock, horse, poultry, equine\end{tabular}                                                                                                                                                                                 \\ \hline
\end{tabular}
\caption{The collected keywords for animal-related vaccine instagram post filtering}
\label{tab:tab-animal-keywords}
\end{table}

\section{Additional Visual Cue Analysis}
\begin{figure}[h!]
\centering
\includegraphics[width=\linewidth]{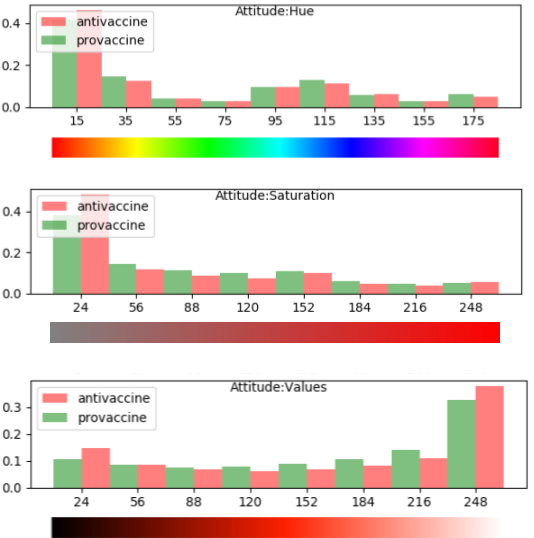}
\caption{Average hue, saturation, and salues histogram comparison across posts with/without anti-vaccine sentiment. The visualization of the values for each channel is placed below the corresponding histogram.}
\label{fig:fig-anti-vaccinehsv}
\end{figure}

\begin{figure}[h!]
\centering
\includegraphics[width=\linewidth]{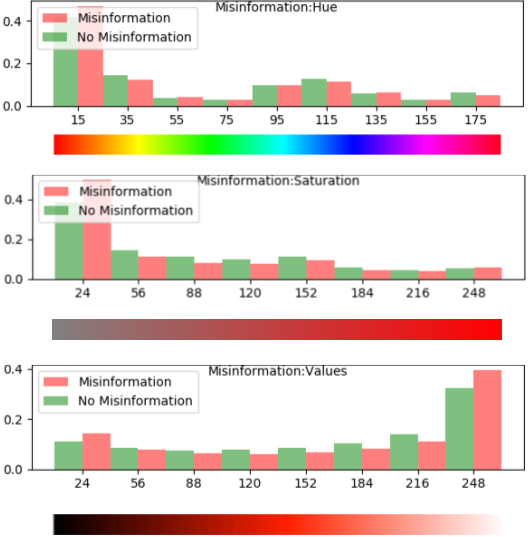}
\caption{Average hue, saturation, and values histogram comparison across posts containing misinformation/no misinformation. The visualization of the values for each channel is placed below the corresponding histogram.}
\label{fig:fig-misinformhsv}
\end{figure}

We have also conducted additional visual cue analysis by examining the usage of the color spectrum. We employ HSV (Hue, Saturation, Value/Brightness) color scheme,  where we compute histograms for the pixel distribution over the three channels from our dataset. Then, we compute average HSV histograms for the group of posts for each class. We compare the average HSV histogram between anti-vaccine posts and non-antivaccine pots. Similarly, we also conduct the comparison between the average HSV histogram of misinformation posts and the ones that does not contain misinformation. The results are summarized in Figure~\ref{fig:fig-anti-vaccinehsv} and Figure~\ref{fig:fig-misinformhsv}

% % We evaluated normalized histograms for the three parameters over all 4997 images, for both . 
\begin{figure*}[!htb]
    \centering
    \includegraphics[trim={0 0 0 0},clip,width=\linewidth]{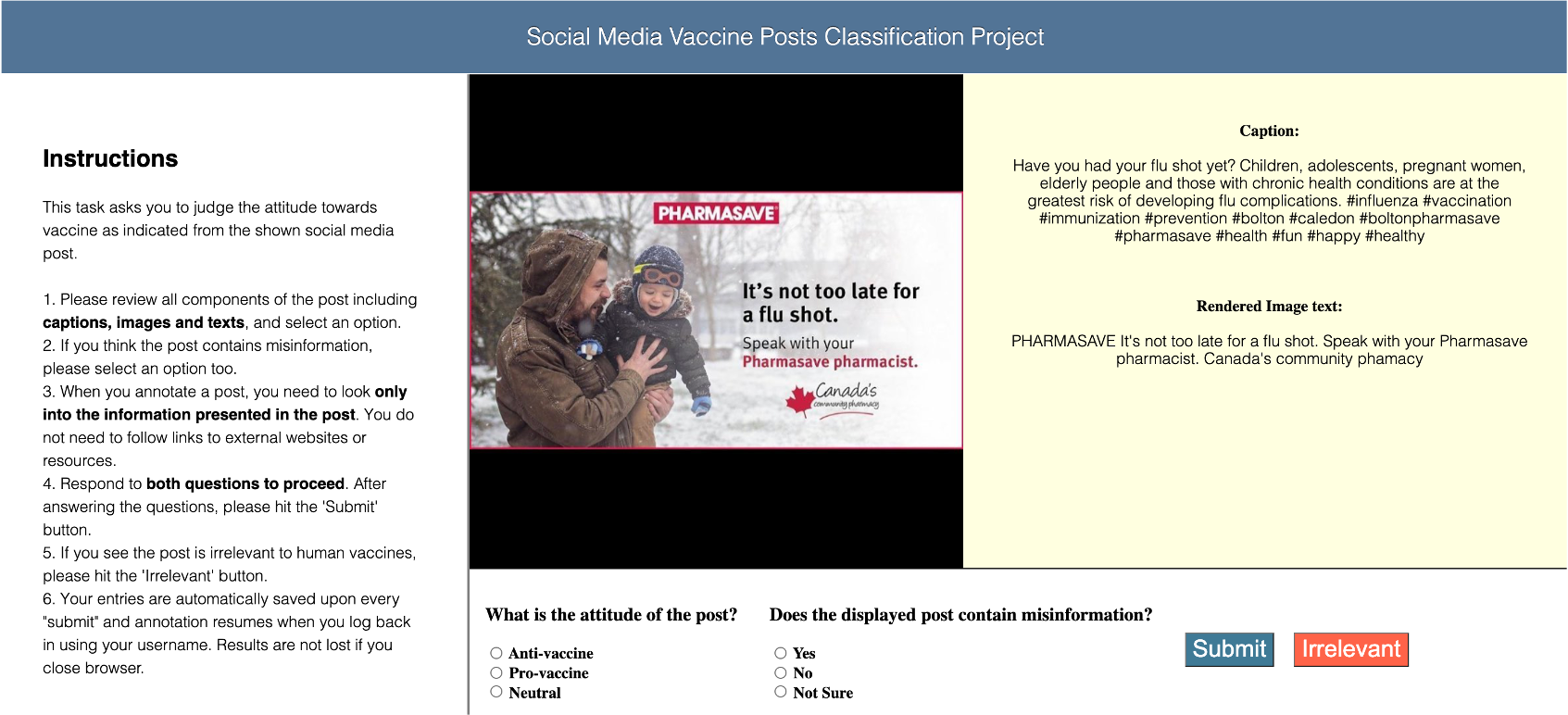}
    \caption{Interface to collect the annotation on anti-vaccine and misinformation post from the Amazon Mechanical Turkes}
    \label{fig:interface_collect_annotation}
\end{figure*}

% When we compare the average histogram of the post with anti-vaccine sentiment against that of the opposite group, we observe a similar distribution of values on each channel. However, if we only compare the dominant values from each channel, we would notice a clear different between the two groups. Regarding the Hue value, we find that anti-vaccine post has more preference on colors with low hue values (such as red). Additionally, anti-vaccine post also contain more colors with low saturation value which indicates that the images would appear to be less vivid than that from that opposite group.  Finally, anti-vaccine users tend to use brighter (high Value) images accompanying their posts. A very similar trend is also sustained when comparing misinformation posts with its opposite group.  

\section{Dataset Annotation Interace} \label{insta-interface}
We demonstrate our data annotation interface in Figure~\ref{fig:interface_collect_annotation}. On the left-hand side, we detailed the instruction to help annotators to accomplish the task. The image and the associated captions as well as text presented in the image are displayed in the middle and on the right-hand side. Finally, at the bottom there is a multiple choice question for annotators to fill in. 

\section{Ethical Statement}
Our dataset and code will be publicly available upon notification of the paper acceptance. We will publish the data in agreement with Instagram's Terms and Conditions \cite{instagram_data_policy}, where we would not directly share the original post content but just distribute the Post ID's. Researchers can then simply retrieve post content through these IDs by using open-source APIs such as Instaloader.

\end{document}